\documentclass[11pt]{article}
\usepackage[margin=1in]{geometry}
\usepackage{times}
\usepackage{amsmath,amssymb}
\usepackage{booktabs}
\usepackage{graphicx}
\usepackage{float}
\usepackage[colorlinks=true,linkcolor=blue,citecolor=blue,urlcolor=blue]{hyperref}
\title{Pre-carved Niches: The Formation Dynamics of Modular Task Partitions in Early LLM Training}

\author{
\begin{tabular}{cc}
Guangqi Li$^{1}$ & Yongxin Li$^{1,\dagger}$\\
qingshishiyi@gmail.com & $\dagger$yxlinuist@163.com
\end{tabular}
\\[3pt]
$^{1}$Zaozhuang University, Zaozhuang, Shandong, China\\
$\dagger$Corresponding author
}

\date{}

\begin{document}

\maketitle

\maketitle

\begin{abstract}
Large language models exhibit a modular internal organization that mirrors well-studied functional networks of the human brain, but how this organization forms during training is unknown: prior work has characterized finished models, not the formation process. We track formation step by step: we train a Pythia-410M model from scratch (two trajectories, bf16 and fp32) and run attribution patching at every step, alongside probes for gradient norms, effective updates, weight norms, and first-order loss decomposition across 14 tasks in four cognitive domains. Three findings. First, the modular map is pre-carved: before any learning, the dominant task pair already overlaps at $\approx$3.6$\times$ the attribution substrate (a task-independent baseline), and its layer-0 concentration is an architecture-level constant on this model family. Second, the partition locks in through two sharp jumps whose amplitudes do not track the learning-rate schedule (the second reaching $20.4\sigma$ quiet-window / $6.2\sigma$ global), accompanied by gradient-level relative deprivation---winners receive 2.25$\to$2.73$\times$ the loser's gradient supply, 9.5--11.5 standard deviations below a random control---that does not propagate to updates or weights. Third, deviation from the substrate appears only in the domain being learned, consistent with the hypothesis that modularity tracks learning. We close by separating the feature-level account we can defend from the mechanistic questions we cannot, and we pre-register the scale-threshold hypothesis behind our ongoing 2.8B experiments.
\end{abstract}

\section{Introduction}

Functional specialization is a hallmark of human cognition: language, formal reasoning, social reasoning, and physical reasoning are each supported by distinct, reproducible cortical networks. Whether such modular organization is a general principle of intelligent systems---or an evolutionary accident of biological brains---has become an empirically testable question with the rise of large language models. Han et al.\ (2026) showed that six instruction-tuned LLMs (24B--123B parameters) exhibit a modular architecture aligned with these four domains: tasks sharing a human brain network recruit overlapping sets of their top-0.1\% most-attributed neurons (mean pairwise Jaccard overlap 12.9\% within-domain vs.\ 3.0\% cross-domain), and replacing one task's top-0.1\% neurons with their alternative-input activations damages same-domain tasks far more than cross-domain ones (drop in both-correct accuracy---the fraction of examples where the model assigns higher likelihood to the correct continuation on both the original and the alternative inputs; chance 0.25, not standard top-1 accuracy---of 25.9\% vs.\ 2.5\%; a 10.3$\times$ asymmetry). Yet the authors themselves leave the central question open: why does this modularity form? Their project page poses the question with no experiments attached.

Existing proposals divide into two camps. The functional-necessity view holds that modularity arises to avoid interference between simultaneously learned computations---a pressure that plausibly operates at training time. The historical-contingency view holds that the final partition is a path-dependent artifact of early training noise, not a convergent optimum. Both camps lack direct evidence, because all prior work has characterized trained networks. The formation process itself---what happens to task-specific neural populations during the earliest steps of training---has not been observed at the resolution at which it unfolds.

We close this gap by tracking module formation step by step. Attribution patching is here a borrowed instrument, not an object of study: Han et al.\ used it to map trained models, and we use it unmodified as a time-resolved instrument to measure formation dynamics. Our contribution is the dynamics it reveals---the pre-carved map, the lock-in jumps, the gradient-layer deprivation---plus the measurement discipline (substrate-calibrated deviation, dual-calibration of significance, multi-probe cross-checks) that converts the instrument's output into dynamics evidence. We train a Pythia-410M model from scratch and run attribution patching (the same minimal-pair localization used by Han et al.) at every training step, together with a multi-probe system that additionally records per-neuron gradient norms, effective updates, weight norms, task-conditioned gradients, and first-order loss decompositions across 14 tasks in four cognitive domains. Our contributions are:

\begin{enumerate}
\item \textbf{The map is pre-carved.} Before training has done anything measurable, the task-overlap structure already exists (dominant pair $\approx$3.6$\times$ above the attribution substrate at step 0), its layer-0 concentration is stable across six corpus variants ($\approx$40\%), suggesting an architecture-level constant of the pipeline on this model family, and a construction-level separation factor is reproducible across seeds---but we explicitly resolve this to the feature level and leave the underlying mechanism open.
\item \textbf{The partition locks in through two sharp jumps, not gradual growth.} Overlap rises in two sharp transitions (steps 18--25 and step $\approx$80), the second being the largest event of the run ($+20.4\sigma$ quiet-window / $6.2\sigma$ global, defined in Section~3); the locked state is dynamically maintained under high roster churn rather than frozen.
\item \textbf{The partition is accompanied by gradient-level relative deprivation that does not propagate.} Winner neurons receive 2.25$\to$2.73$\times$ the loser's gradient supply (sum of per-neuron gradient norms; a ratio 9.5--11.5 standard deviations below a random same-size control), yet effective updates and weight magnitudes show no corresponding asymmetry; and deviation from the substrate appears only in domains whose loss is actually being driven down, establishing a tight coupling between modularity and learning.
\end{enumerate}

We also report eight quantitative falsifications (of turnover, migration, absolute gradient starvation, one-shot irreversibility, a weight-equilibrium model, and three mechanistic factor hypotheses) as negative results that sharply constrain what the formation mechanism can be.

\section{Related Work}

\textbf{Static modularity in LLMs.} Han et al.\ (2026) localized task-supporting units with attribution patching across 46 tasks in four domains and demonstrated domain-selective overlap and ablation effects in six instruction-tuned models. Complementary work finds emergent modular structure in dense pretrained transformers \cite{emergentmoe} and recurring interpretable components across architectures and scales \cite{successorheads}. These studies characterize trained networks; none tracks formation.

\textbf{Developmental dynamics in transformers.} Hoogland et al.\ (2024) divided training into developmental stages using loss-landscape degeneracy and showed that in-context learning emerges through discrete stages \cite{developmental}; circuit-level analyses of grokking \cite{varma2024,nanda2023} established that a generalizing circuit competes with and eventually replaces a memorizing one under weight decay. These works track circuits for single tasks in small models. We extend this developmental lens to multi-task modular partitions in a pretraining-scale model, at per-step resolution.

\textbf{Why modularity emerges.} The interference-avoidance account \cite{han2026,dobs2022} and contingency accounts \cite{achille2019} make opposite predictions about the origin of partitions. Our data bear on both: partitions are largely shaped by input statistics and architecture before learning, and the dynamics that follow are amplification, not competition-driven replacement.

\section{Method}

\textbf{Training.} We train Pythia-410M \cite{pythia} (24 layers, 4096 MLP neurons) from scratch on a pure Pile stream with batch $2^{21}$ tokens per step and a linear learning-rate ramp ($2\times10^{-6}\to9\times10^{-5}$, AdamW). Two trajectories are analyzed: run1 (bf16) and run2 (fp32, seed 1); both use identical architecture and data stream. We record every step: model checkpoints, five per-neuron probe arrays, task-conditioned gradients for 14 tasks, and attribution tensors.

\textbf{Tasks and minimal pairs.} We use 14 tasks spanning four domains: 8 language tasks (including three determiner--noun agreement variants: regular, irregular, and with-adjective), 2 theory-of-mind tasks, and 4 physical-reasoning tasks (temperature, solubility, elasticity, buoyancy), following the battery of Han et al.\ Each task consists of minimal pairs---original and alternative prompts differing at one position such that the correct continuation flips.

\textbf{Attribution patching and rosters.} Following Han et al., a neuron's attribution is (activation difference between original and alternative) $\times$ (gradient of the original--alternative logit difference), aggregated over examples at the final prompt position. A task's roster is the set of its top-983 neurons by signed attribution ($>$0); the roster size $k=983$ equals 1\% of all neurons. The pairwise overlap of two tasks is $|A\cap B|/983$. This single-sided fraction differs from the Jaccard overlap used in prior work (Han et al.): for small overlaps Jaccard $=s/(2-s)$, roughly half the single-sided value (our step-1 dominant-pair overlap 0.140 corresponds to $\approx$0.075 Jaccard), and our top-1\% roster size likewise differs from their top-0.1\%; we never compare raw numbers across these metrics.

\textbf{Deviation from substrate.} Raw overlap mixes task-specific sharing with a domain-specific shallow-layer substrate shared by many tasks regardless of content. We measure the substrate explicitly (attribution substrate $0.0422$, defined as the mean step-1 sibling-pair overlap, where sibling pairs are the non-dominant within-construction pairs regular--adjective and irregular--adjective) and report \textbf{deviation = raw overlap / substrate} throughout. We do not use ``$\times$random'' multiples as overlap effect sizes: the random expectation (1\%) is not the relevant zero point, and conflating the two inflates effect sizes (e.g., 15.26$\times$ random equals $\approx$3.6$\times$ substrate deviation).

\textbf{Multi-probe system.} Per step we record five per-neuron arrays (gradient norm, effective update norm, weight norm, activation frequency, second-moment), task-conditioned gradient scalars for all 14 tasks, and the first-order loss decomposition $\Delta L_i \approx g_i \cdot a_i$ per neuron per task. Each probe serves a distinct evidentiary role and guards against a specific confound (Table~\ref{tab:probes}).

\begin{table}[H]
\centering
\small
\caption{Multi-probe system: each probe establishes one evidentiary link and guards against one confound.}
\label{tab:probes}
\begin{tabular}{@{}l p{0.30\linewidth} p{0.30\linewidth}@{}}
\toprule
\textbf{Probe} & \textbf{Establishes} & \textbf{Guards against} \\
\midrule
Attribution patching & task-specific causal contribution of neurons & correlation-only localization \\
Gradient norm (task-conditioned) & gradient supply to a population & claim of starvation without supply data \\
Effective update norm & whether deprivation reaches optimizer updates & conflating raw gradients with effective steps \\
Weight norm & whether populations materially shrink & ``starved'' claims without weight evidence \\
First-order loss decomposition & which domains are being learned & domain-coupling claims without loss evidence \\
Roster tracking & membership identity, churn, layer position & aggregate curves hiding turnover \\
\bottomrule
\end{tabular}
\end{table}

\textbf{Event detection.} We detect transitions as first differences of the mean overlap series, and report each transition's size in units of $\sigma$: $\sigma$ is the quiet-window dispersion of those first differences, computed as $1.4826\times$ the median absolute deviation of the first-difference series over the pre-event plateau (steps 1--17, the quiet window)---a robust standard-deviation estimate. Because significance depends on this calibration, every event is dual-reported: quiet-window $\sigma$ (0.00754) and the same estimator over the full run (global $\sigma = 0.0249$). The two can disagree strongly---the second jump is $20.4\sigma$ quiet-window but $6.2\sigma$ global---which is why both calibrations are always stated.

\textbf{Caliber discipline.} Two localization calibers exist in our pipeline (signed top-k vs.\ absolute-value top-k); their rosters overlap only 0.28 (Jaccard) and their scores correlate only 0.30 (Spearman), so results are never mixed across calibers. All numbers in the main text use the signed caliber (PRIMARY), frozen in a written protocol (details in Appendix~A).

\begin{figure}[t]
\centering
\includegraphics[width=\linewidth]{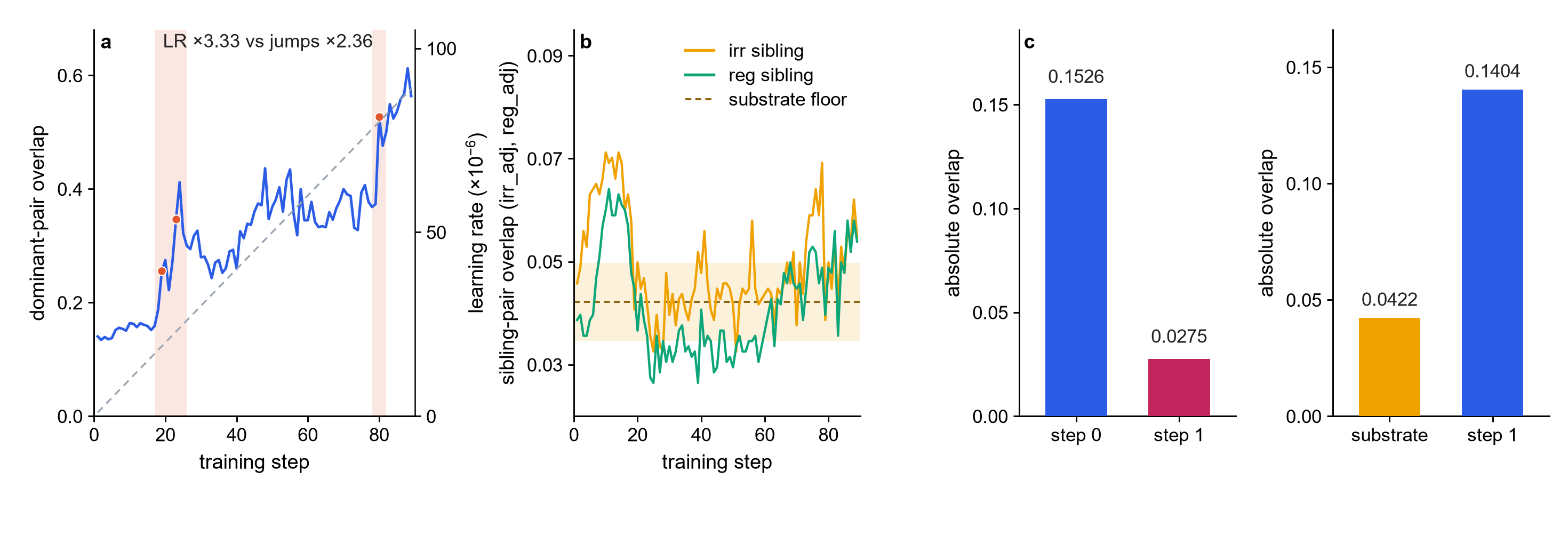}
\caption{\textbf{The pre-carved map and its lock-in} (run2, steps 1--89). (a) Dominant-pair overlap (left axis) vs.\ learning-rate ramp (right axis). (b) Sibling-pair overlaps (irregular--adjective, regular--adjective) with the attribution-substrate floor (dashed). (c) Absolute overlap at steps 0 and 1: dominant pair, step-0 siblings, substrate.}
\label{fig:hero}
\end{figure}

\section{The Map Is Pre-carved}

\subsection{A partition sketch exists before training}
At step 0---with random weights, before any update---attribution patching already recovers the partition sketch (Figure~\ref{fig:hero}): the dominant pair (regular--irregular) shares $0.153$ raw overlap ($\approx$3.6$\times$ substrate), while sibling pairs involving the adjective variant share $0.028$--$0.029$, \emph{below} the substrate level. At step 1 the same structure holds (dominant $0.140$, $\approx$3.3$\times$). The step-1 dominant overlap reproduces across trajectories (0.1404 vs.\ 0.142; difference 0.0016) and across initialization seeds (coefficient of variation 0.036). The pre-carved map is therefore a property of the task structure interacting with the (untrained) network, not of any training history.

\subsection{A layer-0 concentration band that nothing moves}
\begin{figure}[t]
\centering
\includegraphics[width=\linewidth]{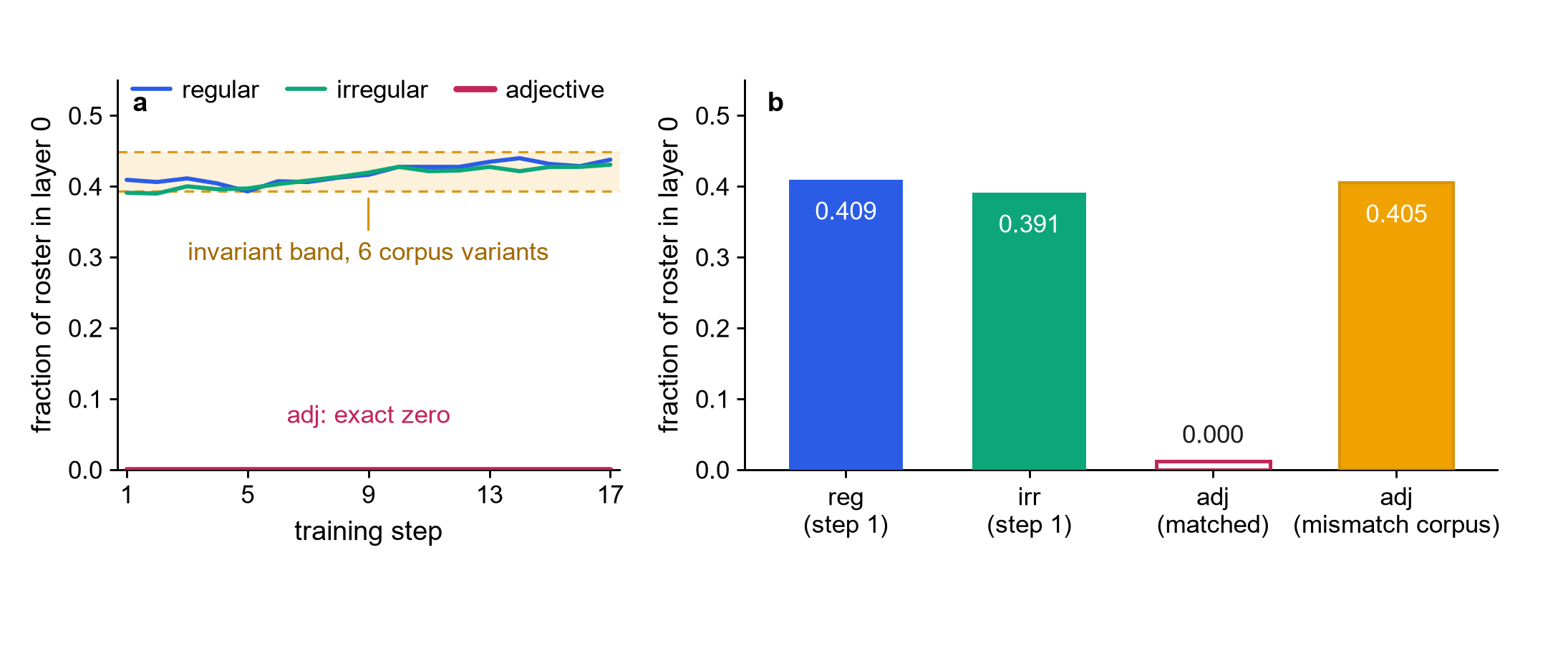}
\caption{\textbf{Layer anatomy.} (a) Fraction of roster in layer 0 over the step 1--17 plateau for the three determiner--noun tasks (regular, irregular, with-adjective). (b) Step-1 layer-0 fraction for regular, irregular, and adjective under matched vs.\ mismatch corpus.}
\label{fig:layers}
\end{figure}

The winner tasks place 39--45\% of their rosters in layer 0, and this fraction is invariant across six corpus variants---synthetic templates, symbol-output answers, random-string bodies, real corpus, before and after training (Figure~\ref{fig:layers}). We take this narrow band as an architecture-level constant of the attribution pipeline on this model family. The adjective variant is the exception: its layer-0 attribution is \emph{exactly zero} (all 98{,}304 neurons), not merely small. Replacing the corpus with a mismatch variant restores it to 0.405, showing the zero is corpus-dependent, not a pipeline boundary. The same variant's roster centroid (mean layer of its roster members) starts at layer 7.9 and decays to 3.7 by step 85; its ``deep residence'' is consistent with the shadow of shallow-layer exclusion rather than a deep-layer attractor (F7).

\subsection{A construction-level separation factor, resolved to features but not to mechanism}
We reverse-engineered the separation by generating faithful-corpus variants (same templates, same answer pools) and minimal variants that isolate single factors. The sibling-pair suppression reproduces across two seeds (raw overlaps 0.024--0.038, all below the 0.042 substrate). What we can defend: within the determiner--noun agreement construction, the structural relation between the critical determiner and the answer position is robustly associated with the binary switch in attribution patterns. What we cannot defend: \emph{which} feature---the adjacency topology, the word class of the intervening word, or position---is the driver. Our two isolation experiments point in opposite directions (a random-word insertion between determiner and noun does not collapse layer 0, while an adjective interruption does), and we register this contradiction as an open problem rather than resolving it by fiat. We make no claim of sufficiency or generality beyond this construction; a 22-task orthogonal family across five feature axes is designed to test generality (Appendix~F).

\section{The Partition Locks In via Two Jumps}

\subsection{A two-jump structure, not gradual growth}
The dominant-pair overlap evolves in stages: a flat pre-event plateau (0.13--0.16, steps 1--17), a first lock-in jump at step 18$\to$19 ($+0.067$; 8.9$\sigma$ quiet-window), a consolidation plateau with oscillations (0.24--0.44; largest single transition $+0.074$ at step 23, 9.8$\sigma$), and a second jump at step 79$\to$80 ($+0.154$; 20.4$\sigma$ quiet-window / 6.2$\sigma$ global-MAD), the largest event of the trajectory, carrying overlap to 0.53--0.61. The two-jump architecture replicates across trajectories: run1's first jump lands at steps 23--25 ($+0.080$; 16$\sigma$ quiet-window, the only calibration archived for run1) and its second at steps 74--75 ($+0.142$; significance not archived). The second jump is not explained by the learning-rate schedule under a correlational ruling---no learning-rate ablation was run: the learning rate grew 3.33$\times$ between the two jumps while the jump amplitude grew only 2.36$\times$ (0.154 vs.\ 0.065), the LR-normalized jump size \emph{declines} (Figure~\ref{fig:hero}a), and the jump magnitude correlates only weakly with the learning rate (Pearson $+0.216$ overall, $+0.021$ at the events). Its trigger therefore remains an open candidate set---an intrinsic second lock-in, a supply-ratio inflection, or a data-order cluster (Appendix~I).

\subsection{The lock is a high-churn reorganization, not stable amplification}
Inside the first lock window the shared set grows by $+71\%$ (run2) / $+121\%$ (run1), but old-member retention is only 60\%/44\%---neither pure expansion nor replacement, under a frozen two-threshold criterion applied identically to both runs (Appendix~E). Across the full window the winner pair shows zero roster-turnover and zero migration cases (0/117 per-step adjudication cases; turnover = step-to-step retention $<$50\%, migration = roster centroid shift $\ge$2 layers, Appendix~E); the single turnover case in the entire grid is the adjective task at step 79 (retention 49.6\%, Appendix~D). Anchoring---the fraction of final winners that already carried positive attribution at step 1---is 79\% (run1) and 53--64\% (run2): the lock amplifies pre-existing tendencies, but the roster itself remains fluid. The locked state is dynamically maintained: after the first jump the overlap retraces to $0.19$ over steps 34--35 (run1) and recovers to $0.30$ without intervention, ruling out a one-shot irreversible transition. Across trajectories, the pre-jump turnover trough consistently attaches to that trajectory's \emph{largest} jump, suggesting a shared causal pattern whose timing is trajectory-dependent.

\begin{figure}[t]
\centering
\includegraphics[width=\linewidth]{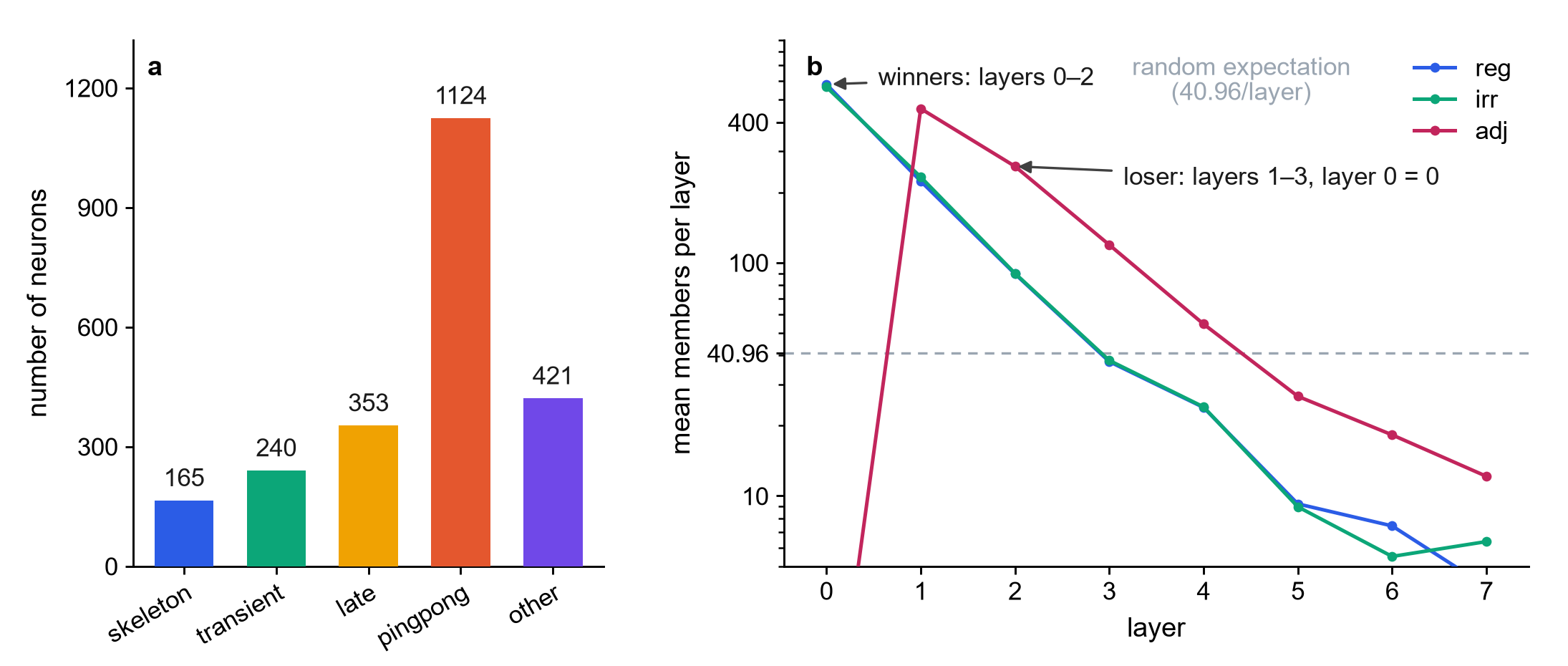}
\caption{\textbf{Core--periphery at neuron resolution.} (a) Fate classes of neurons that ever entered the winner shared set (run2, steps 1--85; attribution-roster caliber). (b) Mean roster members per layer (window 120--165; loss-decomposition caliber): winners (regular, irregular) vs.\ loser (adjective); dashed = random expectation.}
\label{fig:core}
\end{figure}

At the resolution of individual neurons, in run2 the module is a small persistent core inside a large churning periphery (Figure~\ref{fig:core}a): 165 neurons form the skeleton (shared-set members at both step 24 and step 80), and these skeleton members constitute $\ge$90\% of the shared set from step 24 onward (non-monotonically: 56\% at step 35); meanwhile 1{,}124 neurons enter and exit the set at least three times (ping-pong class), 240 are present at step 24 but gone by step 80 (transient), 353 appear at step 80 without having been at step 24 (late recruits), and a residual 421 entered the set without matching any class. The locked state is thus carried by a stable minority of individual neurons with the majority in continuous flux---the neuron-level counterpart of anchoring, and a likely reason aggregate overlap can stay high while the roster keeps moving.

\subsection{Conflict-and-cleaning is trajectory-conditional}
In run1, the shared set's task-gradient agreement collapses below the random-control band during expansion and recovers after pruning (a conflict--cleaning cycle). In run2 this signature is muted: dips are one-third to one-half as deep and never significantly below control, and no large pruning occurs. We report the cycle as a property of run1, not as a general law.

\section{Relative Deprivation That Does Not Propagate, and a Coupling to Learning}

\subsection{Gradient-level deprivation}
\begin{figure}[t]
\centering
\includegraphics[width=\linewidth]{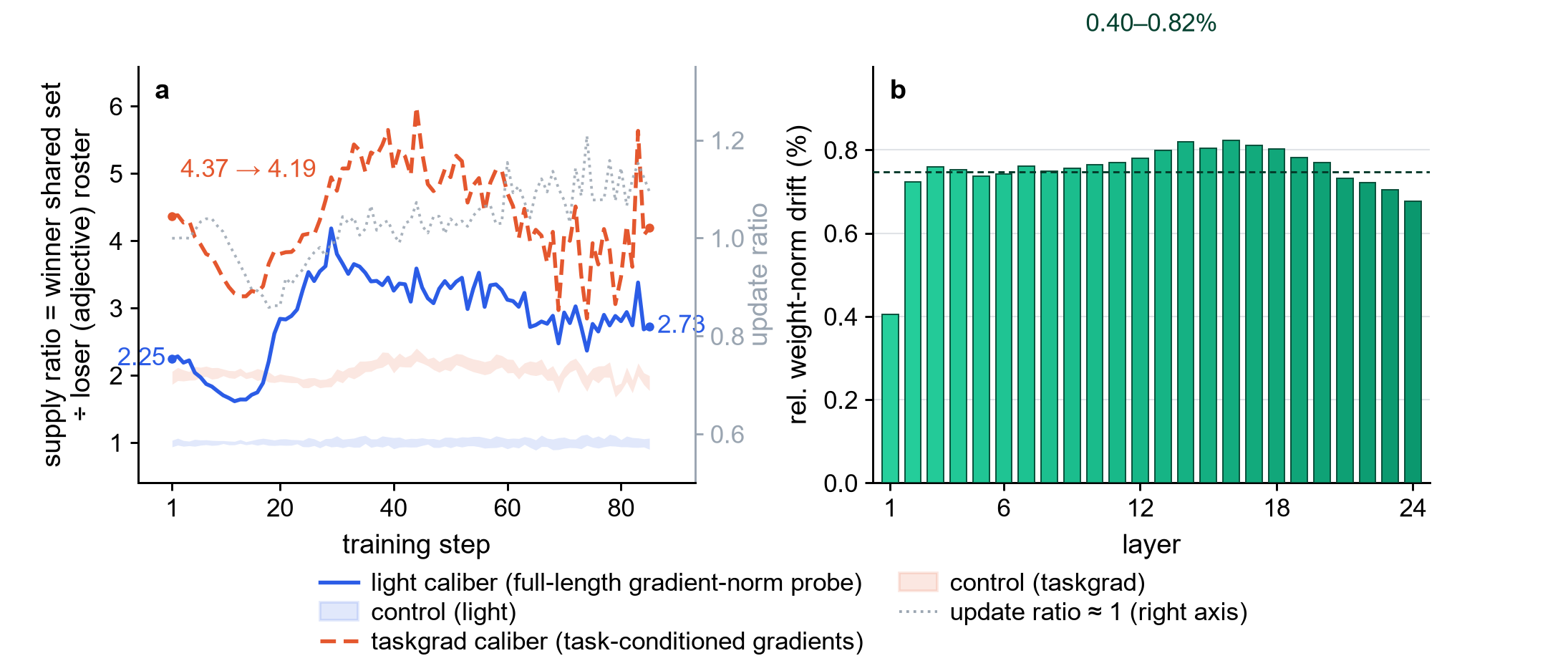}
\caption{\textbf{Gradient-supply profile.} (a) Supply ratio (winner shared set / loser (adjective) roster) vs.\ step, for the light and taskgrad calibers, with random-control bands; dotted gray (right axis): update ratio. (b) Relative weight-norm drift per layer.}
\label{fig:supply}
\end{figure}

Throughout training the winner shared set receives 2.25$\to$2.73$\times$ the loser's gradient supply (light caliber: sum of per-neuron gradient norms over the population, from the full-length gradient-norm probe; the task-conditioned taskgrad caliber gives 4.37$\to$4.19; Figure~\ref{fig:supply}); the loser/winner gradient-norm ratio falls from 0.44 to 0.37, sitting 9.5--11.5$\sigma$ below a random same-size control at every step ($\sigma$ = the standard deviation of the random-control ratio distribution; Appendix~E). This deprivation is confined to the gradient layer: the effective-update ratio (Adam-normalized step sizes) stays $\approx$1, and the weight-norm ratio stays $\approx$1, with weight norms drifting only 0.40--0.82\% over the trajectory. The loser is therefore not materially starved---its exclusion from \emph{learning} is near-total rather than absolute: its first-order loss contribution stays an order of magnitude below the winner's throughout, and late-stage weak learning ($\approx$1/10 magnitude) emerges after step 92 instead of permanent exclusion (Appendix~B, F9), while its parameters remain healthy. The late-stage loser module mirrors the winner's core-periphery structure at higher contrast (Figure~\ref{fig:core}b): it is spatially aggregated (layers 1--3, layer 0 exactly zero in both calibers) and its loss-decomposition working set is stable (adjacent-step Jaccard 0.652, comparable to the winners' 0.690/0.735), yet its attribution roster remains highly fluid (step-165 anchoring 4.8\%) while its working set monotonically converges onto that roster (overlap 0.18$\to$0.40). Which caliber defines module identity is an open question (Appendix~B, F11; Appendix~C). Deprivation is associated with exclusion; the causal direction is not established, and we do not claim starvation as the mechanism of the partition.

\subsection{Deviation occurs only in domains being learned}
\begin{figure}[t]
\centering
\includegraphics[width=\linewidth]{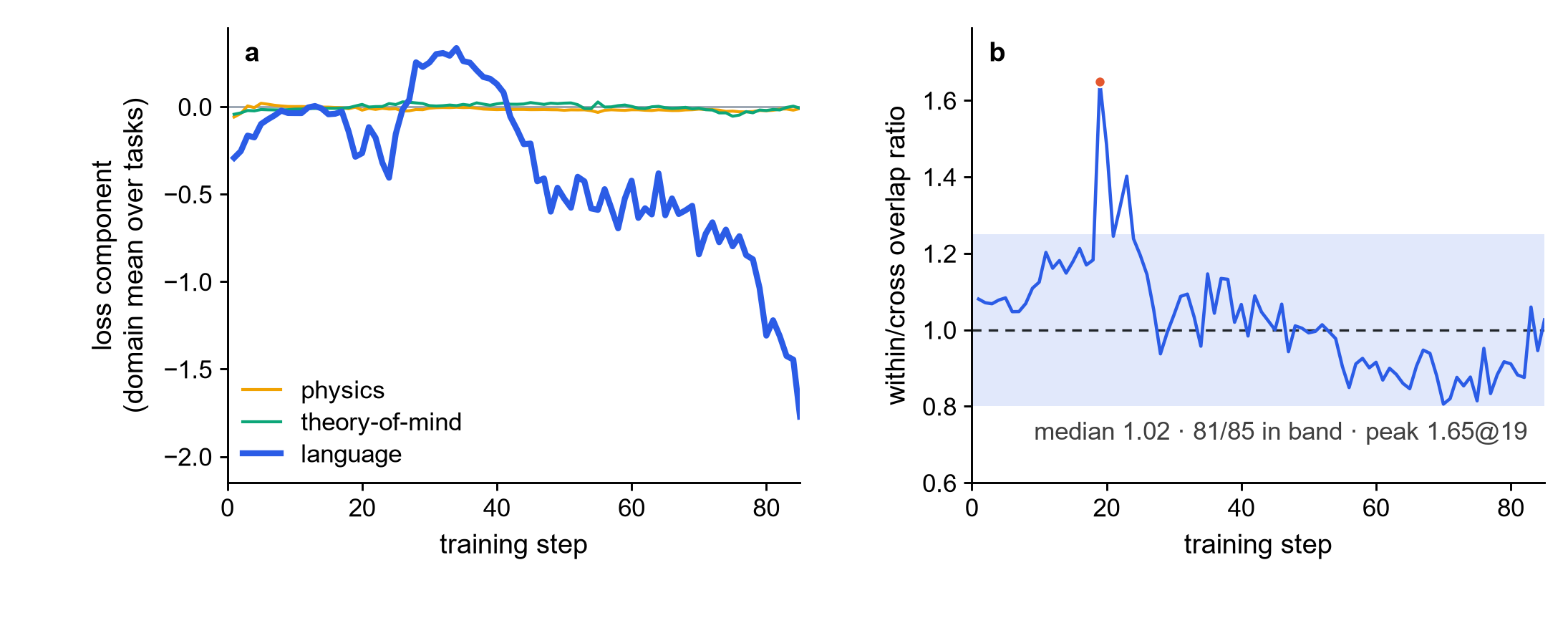}
\caption{\textbf{Deviation occurs only in domains being learned.} (a) First-order loss component by domain (mean over tasks): language, physics, theory-of-mind (ToM). (b) Within/cross overlap ratio (physics).}
\label{fig:domains}
\end{figure}

First-order loss decomposition separates the domains cleanly: the language domain drifts monotonically negative ($-0.29\to-1.77$; irregular deepest at $-1.23$), while physics and theory-of-mind remain structurally flat ($\pm0.05$, no trend), and the adjective task stays near zero (late-stage weak learning at $\approx$1/10 the winner's magnitude; Appendix~B, F9). Both lock-in jumps occur exclusively in the language domain. Consistently, the physics domain's within-domain/cross-domain overlap ratio (mean within-domain overlap $\div$ mean cross-domain overlap) sits at $\approx$1.02 for 81/85 steps inside [0.8, 1.25] (i.e., on the substrate), with a transient two-task burst at steps 19--23 (elasticity--buoyancy, ratio peak 1.65 vs.\ the $\approx$1.02 substrate level) that decays immediately (Figure~\ref{fig:domains}). The refined working hypothesis---deviation from the substrate occurs only in domains being learned---is consistent with all evidence; its scale dependence (physics modularity first appearing at 2.8B---relayed evidence from official checkpoints, Appendix~G---absent at 410M and at the 1.4B checkpoint) motivates the pre-registered threshold experiments in Appendix~G.

\section{Discussion}

\textbf{Three levels of answer.} At the \emph{feature} level, the step-0 map is shaped by three factors: a construction-level structural separation, output word-form as a roster-identity modulator (synthetic-corpus ablations; Appendix~F), and an architecture-level layer-0 concentration band. At the \emph{dynamics} level, partitions lock in via two sharp jumps whose amplitudes do not track the learning-rate schedule, under high churn---a stable skeleton core with a churning periphery at the neuron level---accompanied by gradient-layer deprivation that does not propagate. At the \emph{mechanism} level---why structural separation reshapes attribution alignment in a random network, and where the $\approx$40\% band and the substrate floor come from---we have no answer, and we avoid vocabulary that would imply one: at step 0 there are no circuits to describe, only statistical alignment between input structure, random features, and the output head.

\textbf{Relation to existing accounts.} The interference-avoidance account cannot be the origin: partitions exist before learning and the conflict signature is trajectory-conditional. The contingency account survives only in a weak form: partition topology is highly reproducible across the seeds and trajectories tested; contingency lives in roster identity and event timing.

\textbf{Falsifications as contributions.} We report eight quantitative falsifications that constrain future theory: no roster turnover or migration (0/117), no absolute gradient cutoff (loser supply never below 0.59 of the random-control baseline; cf.\ gradient starvation \cite{pezeshki2021}), no one-shot irreversibility (spontaneous recovery), no weight-equilibrium relation between supply and norms (Pearson $r=-0.27$; Appendix~B, F5), and three mechanistic factor hypotheses rejected (layer-0 local-bigram reading; span-to-depth mapping; noun-pool coherent accumulation).

\section{Limitations}

\begin{itemize}
\item[(i)] The feature-locking result (Section~4.3) is established within one construction family (determiner--noun agreement) with three template variants; generality to other constructions and domains is untested and explicitly open.
\item[(ii)] Dynamics claims rest on two trajectories of a single 410M model; cross-seed training replication is planned but incomplete.
\item[(iii)] The two trajectories differ in numerical precision (bf16 vs.\ fp32) and other factors; the source of their differences (conflict-cycle depth, event significance regime, roster fluidity) is unresolved.
\item[(iv)] Relative deprivation is correlational; no intervention (supply equalization) has yet established its causal role.
\item[(v)] The physics domain at 410M is not learned, so scale-threshold conclusions are pre-registered predictions, not measurements.
\item[(vi)] Two localization calibers are non-interchangeable; conclusions are caliber-specific by protocol.
\end{itemize}

\section*{Reproducibility Statement}
All probe and attribution data are archived per step with SHA-256 bindings (list in Appendix~H); all analysis scripts are stored alongside results; frozen criteria (substrate definitions, deviation formula, event-detection calibrations, adjudication thresholds) are recorded in protocol documents referenced throughout.

\section*{Ethics Statement}
No human subjects or human-derived data beyond the public Pile corpus and publicly released task batteries are used. The experiments use a 410M-parameter model trained on public data.

\appendix


\section*{Appendix A: Measurement Discipline for Formation Dynamics}
\label{app:caliber}

\emph{Question answered: why do formation-dynamics claims require substrate-calibrated deviation and frozen localization calibers, and what happens when they are mixed?} Per-step attribution patching opens two artifact classes that trained-model studies do not face: a mechanical random baseline that inflates effect sizes, and caliber instability across steps. This appendix records the discipline that closes them.

The substrate concept evolved through three stages in this project.
(i) A \emph{domain-blind} substrate (one cross-task baseline for everything) was falsified by analysis A4: all 12/12 units tested are domain-specific---cross-domain roster Jaccards are 0.0596$\to$0.1781 (syntax$|$physics), 0.0182$\to$0.0644 (syntax$|$ToM), and 0.0294$\to$0.0856 (physics$|$ToM); none falls in the 30--60\% mixed band, and the three-domain shared core collapses 0$\to$29.
(ii) A \emph{domain-specific but step-calibrated} substrate followed: analysis A1 showed the step-constant substrate is unstable (adjacent-step Jaccard mean 0.487, min 0.1757), so the substrate is measured per step where used.
(iii) The frozen formula is \textbf{deviation = raw overlap / substrate} (PRIMARY attribution substrate $0.042218$, the mean step-1 sibling-pair overlap). Worked examples: step-1 dominant overlap 0.140387 $\to$ deviation 3.3253$\times$; step-85 dominant 0.536114 $\to$ 12.6987$\times$.

We do not use ``$\times$random'' multiples: the random expectation (1\%) is not the relevant zero point, and conflating the two inflates effect sizes---15.26$\times$ random equals $\approx$3.6$\times$ substrate deviation.

Two attribution calibers exist in the pipeline (signed top-983 with attribution $>$0, PRIMARY; and absolute-value top-983, SECONDARY). They are non-interchangeable: roster Jaccard is at most 0.2838 (regular task, step 85), full-model $|\text{grad}|$ vs.\ $|\text{attr}|$ Spearman is at most 0.2952, layer-wise Spearman medians are $-0.0005$ to 0.0700 (max 0.2406), and step-0/step-85 roster intersections are 483/501/492 and 572/596/484---all below $2/3$ of 983. All main-text numbers use PRIMARY, frozen in a written protocol; calibers are never mixed.

A third, gradient-roster caliber question remains open (Appendix~B, F10): the orderings by $|\mathrm{grad\_norm\_task}|$ and by $|\mathrm{grad\_task}|$ are not equivalent (per-neuron Pearson 0.59--0.71; top-983 sets differ). Gradient-roster conclusions carry their caliber labels until a unique caliber is adjudicated.

\section*{Appendix B: Falsification Cards}
\label{app:falsifications}

\emph{Question answered: which candidate mechanisms does the evidence quantitatively exclude, and what survives each exclusion?}

The main text counts eight quantitative falsifications (F1--F8). Three further cards (F9--F11) come from the extended step-86--165 window and post-date that count.

\begin{table}[h]
\centering
\small
\begin{tabular}{lp{3.0in}l}
\toprule
\textbf{Card} & \textbf{Hypothesis / strongest measured evidence} & \textbf{Verdict} \\
\midrule
F1 & Turnover: the lock replaces the roster. Lowest retention 72.7\% $\gg$ 50\% threshold; 70--97 times the random same-size baseline (1.0\%); 0/117 turnover cases. & excluded \\
F2 & Migration: the locked circuit moves across layers. Centroid layer 0.51--1.16 throughout; shallow fraction $\ge$96.3\%; 0/117 migration cases. & excluded \\
F3 & Absolute gradient cutoff: losers are starved to zero. Loser supply/baseline $\ge$0.59 throughout; shallow layers rise after events (layer 1: 1.60$\to$2.28). & excluded \\
F4 & One-shot irreversible transition. Retrace to 0.19 over steps 34--35 (run1) with spontaneous recovery to 0.30; the lock is dynamically maintained. & excluded \\
F5 & Weight-equilibrium model (TH1): $\|w\|^2=S/\lambda$ predicts (weight-norm ratio)$^2$ = supply ratio. Measured $(w_W/w_L)^2\approx1$ (1.012$\to$0.996) vs.\ supply 2.25--3.26, corr $=-0.27$; earlier ``ratio matches'' were caliber coincidences or definitional identities and are void. & rejected \\
F6 & Factor 2 (layer-0 reads a local determiner--noun bigram). Insertion surgery (random word / punctuation breaking the adjacency) leaves layer-0 membership 428$\to$409 and 410 in the two surgeries (mean 409.5)---no collapse. & excluded \\
F7 & Factor 3 (receptive-field--depth mapping). Span 0--4 centroids 1.91--2.16 flat (no monotonicity); the parse curve cannot explain the adjective variant's 7.9 deep residence. & excluded \\
F8 & Factor 1 (noun-pool coherent accumulation). Sibling pairs stay flat (raw overlap $\approx$0.145, i.e., $\approx$3.4$\times$ substrate; hyperbolic fit $R^2\le0.21$); reg$\sim$adj decreases monotonically; 10/10 series go down first$\to$last---decay saturation, not accumulation saturation. & excluded \\
\bottomrule
\end{tabular}
\caption{Falsification cards F1--F8 (main text ``eight quantitative falsifications'').}
\end{table}

\noindent\textbf{Extended-window cards.}
\textbf{F9} (permanent exclusion of the loser) \emph{excluded}: the adjective task's first-order loss contribution was recorded as permanently positive; re-measurement (G3 + independent audit) finds it negative at step 1 ($-0.037$) and in 69/80 steps of the 86--165 window (first negative at step 92, min $-0.34$ at step 144). Revised: the loser shows \emph{late-stage weak learning} at $\approx$1/10 the winner's magnitude---deprived, not permanently frozen.
\textbf{F10} (gradient-roster ordering equivalence) \emph{excluded}: per-neuron Pearson between the two orderings is 0.59--0.71; the top-983 sets differ, so physics-anchoring conclusions carry caliber labels pending adjudication.
\textbf{F11} (distributed takeover of loser learning) \emph{excluded} (pre-registered): all three predictions failed (P1 0.849 vs.\ random band [0.142, 0.165]; P2 adj$\to$adj overlap 0.282 vs.\ a random same-size expectation of $\approx$0.01; P3 0.652 vs.\ reference 0.0054). The loser instead forms its own weak module at layers 1--3 (layer 0 exactly zero in both calibers). Scope: 410M run2, adjective task, one construction, one scale, loss caliber.

\section*{Appendix C: Anomaly Registry (Paper-Relevant)}
\label{app:anomalies}

\emph{Question answered: which unresolved observations bear on the paper's claims, and what is their disposition?} Source: ledger 03\_anomalies (append-only; full registry lives there).

\begin{itemize}
\item \textbf{A12}: adjective layer-0 total absence (all 98{,}304 neurons)---disposition: phenomenon, corpus-dependent (mismatch corpus restores 405/395 members); dual-caliber confirmed (attribution + loss-decomposition rosters).
\item \textbf{R2-B1}: cross-trajectory random-control relation reverses (run1 significantly below control, run2 not)---disposition: unresolved; feeds the trajectory-identity question (bf16 vs.\ fp32; seed/LR/batch identity unconfirmed).
\item \textbf{A-0165-1}: giant V spike pair at steps 114$\to$116 ($-0.151$/$+0.149$; 19.8$\sigma$ quiet-window)---unattributed; amplitude-saturated $\sim$0.15.
\item \textbf{A-0165-7/F10}: physics roster caliber contradiction (protocol says $|\mathrm{grad\_norm\_task}|$, archive holds $|\mathrm{grad\_task}|$; orderings not equivalent)---awaiting judge ruling on a unique caliber; physics ``entry'' conclusions are caliber-conditional until then.
\item \textbf{A-0165-9}: three-caliber physics entry points diverge (signed $\sim$19 / $|\mathrm{grad\_task}|$ $\sim$91 / $|\mathrm{grad\_norm\_task}|$ almost never).
\item \textbf{R2-108-1}: gradient-cosine 0.93 holds only in the $|\mathrm{grad\_task}|$ amplitude-profile, max-pair caliber (0.927 at step 108); signed mean-pair cosine peaks at step 99 (0.176/0.196) and is $\approx$0 at step 108.
\item \textbf{R2-108-2}: irregular retention at the step-108 trough: 49.34\% (recompute) vs.\ 57.0\% (G4)---definition/caliber discrepancy under review.
\item \textbf{R2-108-3/4, A-0165-5}: step-108 loss spike located as data-driven (same-shaped spikes in run1 at shared steps); residual mechanism questions (domain-selective propagation, energy concentration to layer 0) remain.
\item \textbf{R2-H1-1}: dual-caliber identity split for the adjective task (attribution roster highly fluid---step-165 anchoring 4.8\%---while the loss-decomposition ``working set'' is stable---window Jaccard 0.652---and converges onto the attribution roster)---awaiting judge ruling on which caliber defines module identity.
\item \textbf{A-0165-6}: ``pinned to layers 0--7'' is strictly false (shared-set max layer 3--19, step 1 already 8); valid only as a mass characterization ($>$98\% of members)---display caliber corrected in the main text.
\end{itemize}

\section*{Appendix D: Two-Trajectory Details and Probe Repair History}
\label{app:trajectories}

\emph{Question answered: what exactly distinguishes the two trajectories, and which cross-trajectory statements does the paper make?}

run1 (bf16) and run2 (fp32, seed 1) share architecture and data stream; whether precision is their \emph{only} difference (seed, LR schedule, batch) is unconfirmed, so cross-trajectory differences are not cleanly attributable (ledger 02\_evidence L107; 03\_anomalies L56). Cross-trajectory statements used in the main text:
\begin{itemize}
\item \textbf{Conflict--cleaning is trajectory-conditional} (Section~5.3): run1 shows a full cycle (expansion, below-control collapse, pruning, recovery); run2's dips are one-third as deep ($-0.041$/$-0.059$ vs.\ run1's $-0.114$), never significantly below the random-control band, and followed by alignment without pruning (cosine $+0.319\to+0.622$ at jump 80).
\item \textbf{Turnover/migration}: run1 0/117 both; run2 adds exactly one turnover case (adjective task, step 79, retention 49.6\%), zero migration.
\item \textbf{Anchoring} (signed caliber): 79\% (run1); 0.527/0.643/0.624, i.e., 53--64\% (run2). The membership caliber (step-85 roster $\cap$ step-1 roster) is much lower (15.5/21.8/9.5\%) and answers a different question (identity retention, not sign).
\item \textbf{Layer pinning}: run1 $>$99\% of the dominant pair in layers 0--7 (centroid $\approx$0.5); run2 layer-0 counts grow 90$\to$365 with final centroids 0.54 (regular) / 0.94 (irregular).
\item \textbf{Jump timing}: first jump steps 23--25 ($+0.080$; 16$\sigma$) and 74--75 ($+0.142$) in run1 vs.\ 18$\to$19 and 79$\to$80 ($+0.154$) in run2.
\item \textbf{A/B slot artifact}: the run2 per-step series carries one measurement per step with alternating attribution slots (odd steps slot B, even steps slot A); the resulting 2-step autocorrelation period is a slot-parity artifact, not dynamics. The main-text jump analysis uses the per-step series as-is; significance is computed on its first differences with the slot artifact acknowledged rather than removed.
\end{itemize}

Probe repairs and criterion freezes are recorded in the per-analysis reports (e.g., report\_D for the supply/norm probes) and protocol documents; two audit files (audit\_h1/h2\_p0 and audit\_h5/h6\_p1) are referenced by the reports but missing from the project directory---SI audit provenance for those sections is therefore incomplete and flagged (synthesis \S4.5).

\section*{Appendix E: Statistical Calibrations}
\label{app:calibrations}

\emph{Question answered: how are event significance, anchoring, and adjudication thresholds defined, and why are $\sigma$ values always dual-calibrated?}

\textbf{Event detection.} Transitions are first differences of the mean dominant-pair overlap. Every transition's size is reported in $\sigma$ units, where $\sigma$ is the robust standard-deviation estimate $1.4826\times$MAD of the first-difference series, computed either over the pre-event plateau (steps 1--17, the \emph{quiet window}, giving $\sigma=0.0075412$) or over the full run (the \emph{global} calibration, giving $\sigma=0.0249$). The two calibrations disagree strongly: for run2 the second jump is $+0.154$ with quiet-window $\sigma=0.0075412$ ($20.4\sigma$) but global $\sigma=0.0249$ ($6.2\sigma$). Every $\sigma$ in this paper carries its calibration label; single-calibration significance was an earlier trap this project hit and fixed.

\begin{figure}[h]
\centering
\includegraphics[width=\linewidth]{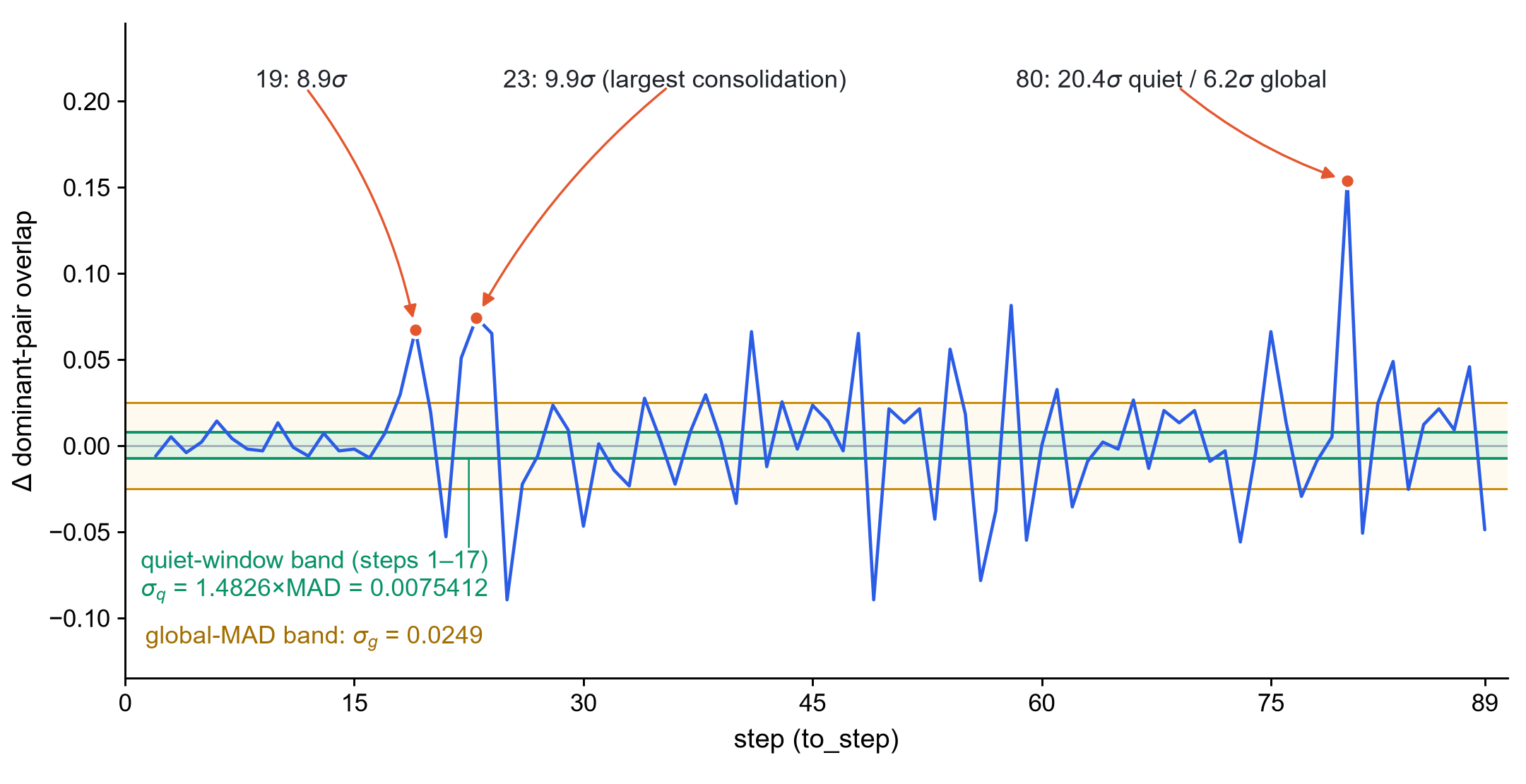}
\caption{First differences of the dominant-pair overlap (run2) with the quiet-window and global-MAD $\sigma$-calibration bands; the marked transitions are the first lock-in jump, the largest consolidation transition, and the second jump.}
\label{fig:sigma}
\end{figure}

\begin{figure}[h]
\centering
\includegraphics[width=\linewidth]{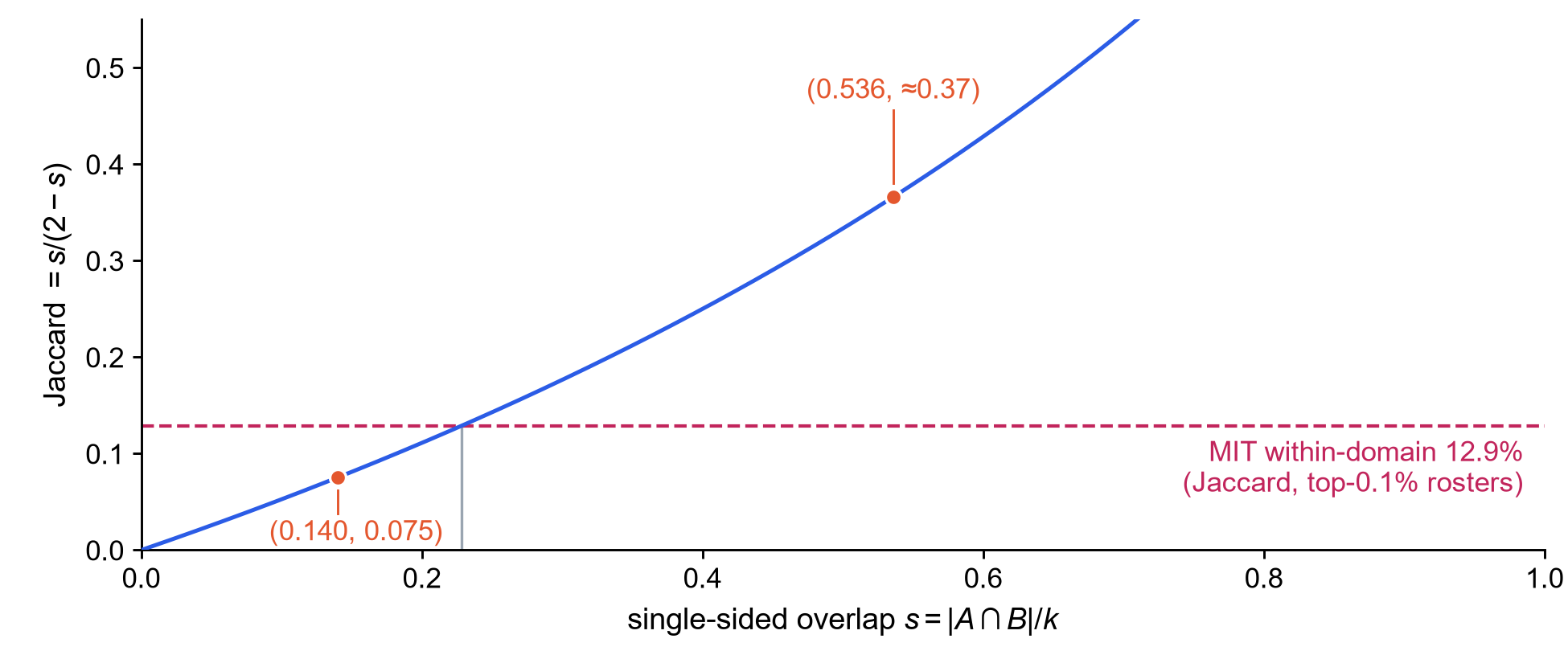}
\caption{Single-sided vs.\ Jaccard overlap conversion. Red points: our step-1 and step-85 dominant-pair overlaps (top-1\% rosters); dashed line: MIT's within-domain reference (top-0.1\% rosters).}
\label{fig:jaccard}
\end{figure}

\textbf{Single-sided overlap vs.\ Jaccard.} Our pairwise overlap is the single-sided fraction $s=|A\cap B|/k$ ($k=983$); the Jaccard index used by Han et al.\ is $s/(2-s)$. For small overlaps the Jaccard value is roughly half the single-sided value (Figure~\ref{fig:jaccard}); our step-1 and step-85 dominant-pair overlaps (0.140 and 0.536 single-sided) correspond to $\approx$0.075 and $\approx$0.37 Jaccard. Raw numbers are never compared across the two metrics, and the roster sizes differ as well (our top-1\% vs.\ their top-0.1\%).

\textbf{Anchoring calibers.} ``Anchoring'' (final winners carrying positive signed attribution at step 1) and ``membership retention'' (step-$t$ roster $\cap$ step-1 roster) are distinct quantities; the paper uses the signed caliber for anchoring (79\% / 53--64\%) and reports the membership caliber separately in Appendix~D.

\textbf{Adjudication thresholds.} Turnover and migration are adjudicated on a grid of 117 per-step cases with frozen thresholds (turnover: step-to-step roster retention $<$50\%; migration: roster centroid shift $\ge$2 layers), applied identically to both trajectories (producing 0/117 for the winner pair, the single adjective-task turnover case at step 79, and zero migration); the deviation formula, substrate definition, and roster size ($k=\lfloor N_{\mathrm{MLP}}/100\rfloor=983$) are frozen in the protocol. The fate classes behind Section~5.2 use fixed anchors: skeleton = shared-set members at both step 24 and step 80; transient = present at step 24 but gone by step 80; late = present at step 80 but not at step 24; ping-pong = at least three membership changes over steps 1--85; other = appeared in the shared set without matching any of the above. The class definitions overlap by construction (a neuron may satisfy several); ``other'' is the complement of the union, and the five counts cover the 2{,}303 distinct neurons that ever entered the shared set.

\textbf{Supply and norm significance.} At each step the loser/winner gradient-norm ratio is compared against random same-size rosters; $\sigma$ is the standard deviation of that random-control ratio distribution (the control mean and SD columns of the archived norm-ratio table), and the measured ratio sits 9.5--11.5$\sigma$ below the control mean at every step.

\section*{Appendix F: 22-Task Orthogonal Family (Design Frozen, Not Yet Run)}
\label{app:family}

\emph{Question answered: how will the construction-level feature separation (Section~4.3) be generalized beyond one construction, and what is the regression that will test it?}

Three tasks (one dominant pair plus two sibling pairs) cannot separate feature axes: regular/irregular share everything, and the adjective variant differs by exactly one structural feature, so template similarity, output word-form, and dependency span cannot be disentangled. The 22-task orthogonal family is the answerable form of the root-cause question (R-P1-1). It is \emph{design-frozen and hash-bound} (orthogonal\_task\_family\_design.md) but not yet executed; the frozen v2 fp32 checkpoints allow back-filling attribution at steps 0/1/25/50/85 without retraining.

Five feature axes: axis 1, template similarity (REG/IRR/D1/D2/D3, $d=0$--3, span pinned to 0); axis 2, output word-form (VERB/NOUN/DEM/AUX); axis 3, dependency span (S0--S3 adjective ladder plus PP/RC subject--verb versions); axis 4, answer vocabulary (V100/V60/V20/V0); axis 5, semantic pairs (HYPER/CATEG/FIELD/SYN/ANT). The frozen 22-row feature$\times$task matrix is transcribed in the design document.

Word-form sensitivity on axis 2 is already measurable in synthetic corpora: swapping the answer token replaces $\approx$85\% of roster members (synthetic answer-token variants reach overlaps 0.910/0.887/0.897, $\approx$21$\times$ substrate, against 0.135--0.229 for the base answer pool). These synthetic numbers do not transfer quantitatively to the real battery (real sibling-pair suppression is 3.7--4.5$\times$, far below the synthetic 9--24$\times$ floor), which is why axis 2 tests word-form effects inside the real-task family rather than in synthetic probes.

\textbf{Regression specification} (pre-registered, not run): step-0 attribution-overlap statistics regressed on the five axes via a response-surface model with coefficients $\beta_1$--$\beta_6$; the coefficient pattern on axis 3 (span) vs.\ axis 2 (word-form) is the direct test of the unresolved R-P1-1 contradiction (random-word insertion does not collapse layer 0 while an adjective interruption does).

\section*{Appendix G: Cross-Scale Protocol and the 2.8B Pre-registration}
\label{app:scale}

\emph{Question answered: what is the pre-registered test of the scale-threshold hypothesis, and which of its parts have measurements behind them?}

\textbf{S1--S6 signatures} (frozen, three tiers): S1, pre-carved ordering (dominant deviation strictly highest; dominant/sibling $\ge$1.7); S2, layer-0 signature (dominant layer-0 $>$0 and adjective layer-0 $<k/100$); S3, layer pinning (dominant $d<0.15$ and adjective $d>0.25$); S4, lock morphology ($\ge$1 event $\ge5\sigma$ under the quiet-window calibration and non-transient); S5, supply ratio (winner/loser $\ge$2.0 and $\ge5\sigma$ under the random-control supply calibration); S6, deviation follows learning (learned-domain deviation rises monotonically; unlearned domain stays $\approx$1 flat).

\textbf{Physics four-branch criterion} (frozen): the threshold is crossed iff (i) learning starts (loss\_end/loss\_start $<$0.90 and monotone), (ii) deviation appears ($>$1.5 sustained for $\ge$5 steps without falling back to $\le$1.2), and (iii) deviation lags learning. Branches: \textcircled{1} capable-but-not-modular (new finding, M-c); \textcircled{2} deviation without learning (anomaly escalation); \textcircled{3} neither (conflict review); \textcircled{4} all three (threshold crossed). Depth prediction: $d_{\mathrm{physics}}>d_{\mathrm{grammar}}$ (equality or less is an anomaly).

\textbf{410M physics evidence (measured)}: loss\_task step 1$\to$85 is 5.545$\to$5.142 (net $-0.40$, near-flat); loss-decomposition $|$mean$|$ collapses $\sim$3e-5$\to\sim$3e-6 ($\sim$10$\times$); the within/cross overlap ratio median is 1.0199 with 81/85 steps in the band [0.8, 1.25]; a transient elasticity--buoyancy burst (steps 19--23; pair overlap 0.395--0.499, ratio peak 1.65) decays immediately. The scale-threshold lower bound is (1.4B, 2.8B] (physics modularity absent at 410M and 1.4B, present at 2.8B per official checkpoints---relayed evidence, awaiting official interpolation).

\textbf{2.8B pre-registration (predictions, no measurements yet)}: the working hypothesis is module threshold = learning threshold $\times$ ``deviation follows learning'', with three candidate hypotheses (H-scale-1 ICL threshold; H-scale-2 domain-level deprivation; H-scale-3 capacity/landscape). The protocol pre-registers an early self-training run of 50 steps measuring S4/S5/S6, lock morphology, and layer structure, under the unified gate 0--4 system---design gate, then the 410M gate, then the generalization gate (S1--S3 first judgment), then the long-training gate, then the physics-entry gate (four-branch). Engineering feasibility (estimated, explicitly labeled): 2.8B $\approx$ 6.85$\times$ 410M; step-0 attribution $<$1 min per 14 tasks; 100M$\to$104.9M tokens $\approx$ 18--29 h. The third act therefore has measurements on one side (410M physics not learned) and frozen criteria with zero measurements on the other (2.8B entry dynamics)---the largest evidence gap of this paper, stated rather than hidden.

\section*{Appendix H: Data and Reproducibility}
\label{app:data}

\emph{Question answered: where are the artifacts behind every number in this paper, and how are they bound?}

Per-step model checkpoints, five per-neuron probe arrays, task-conditioned gradients for 14 tasks, and attribution tensors are archived per step. Every artifact that enters a claim is SHA-256-bound in the project's hash ledger (ledgers/05\_hashes.md in the project directory), which lists the bound result files (analysis reports, CSVs, npz archives, protocol documents, and this paper's PDFs); analysis scripts are stored alongside their outputs, and all frozen criteria (substrate definitions, deviation formula, event-detection calibrations, adjudication thresholds) live in protocol documents referenced throughout. Two provenance gaps are flagged rather than concealed: the run1 40-step full tables are archived on the training server but not yet copied into the project directory (and therefore unbound), and the audit files audit\_h1/h2\_p0 and audit\_h5/h6\_p1 are referenced by reports but missing from the project directory.

\section*{Appendix I: Future Work}
\label{app:future}

\emph{Question answered: which open hypotheses have pre-registered or planned experiments behind them?}

\textbf{Second-jump hypothesis (unfrozen).} Three candidates explain the second jump's trigger: (a) an intrinsic second lock-in via threshold accumulation; (b) a supply-ratio inflection; (c) a data-order cluster. A pre-registered prediction is required before adjudication (e.g., whether the supply ratio shows an inflection or plateau before the second jump, and whether its timing aligns with a data-order cluster boundary).

\textbf{Word-frequency axis (zero-cost, not done).} If relative deprivation approximately equals the corpus-frequency ratio of the three constructions, the mechanism is data statistics rather than competition---the ``competition narrative vs.\ data narrative'' watershed (A16/M4).

\textbf{Answer-position axis (B1 ablations).} Three targeted variants separate word-class sensitivity from adjacency topology: sentence-initial corruption; a multi-word adjective phrase replacing the single adjective; and official-template words replaced by random words---the direct test of the R-P1-1 contradiction (random-word insertion does not collapse layer 0 while adjective interruption does).

\textbf{22-task family execution} (Appendix~F): back-filled attribution at steps 0/1/25/50/85, then the $\beta_1$--$\beta_6$ regression; the sole path off the single-construction limitation.

\textbf{2.8B run} (Appendix~G): gate 2 plus GPU release, then the 50-step early run and the four-branch physics adjudication.

\textbf{H-loser-2}: re-testing distributed takeover for losers at 2.8B across multiple constructions (F11's scope is one task, one construction, one scale).

\end{document}